\documentclass{article}

\usepackage{ai_lab_2026}

\usepackage[utf8]{inputenc} 
\usepackage[T1]{fontenc}    
\usepackage[pagebackref=true,breaklinks=true,letterpaper=true,colorlinks,bookmarks=false]{hyperref}
\usepackage{url}            
\usepackage{booktabs}       
\usepackage{amsfonts}       
\usepackage{nicefrac}       
\usepackage{microtype}      
\usepackage{xcolor}         
\usepackage{amsmath}
\usepackage{multirow,multicol}
\usepackage{graphicx}
\usepackage{makecell}
\usepackage{subcaption}

\usepackage[table]{xcolor}
\definecolor{ngreen}{HTML}{03C75A}
\definecolor{ngreenlight}{HTML}{E8F8EF}

\usepackage{tcolorbox}
\usepackage{microtype}
\usepackage{cancel}

\usepackage{xspace}
\newcommand{\ie}{\textit{i.e.,}\xspace}
\newcommand{\eg}{\textit{e.g.,}\xspace}

\title{On-Policy Delta Distillation}

\author{
    Byeongho Heo
    \quad
    Jaehui Hwang
    \quad
    Sangdoo Yun
    \quad
    Dongyoon Han
    \\ [2mm]
    NAVER AI Lab
}

\begin{document}

\maketitle

\begin{abstract}
On-policy distillation is an alternative post-training method in reinforcement learning that alleviates the constraints imposed by reward models by providing token-level supervision from a teacher model. Although on-policy distillation has been studied and applied across various settings, its fundamental design remains underexplored. In this paper, we introduce a new distillation reward, termed \textit{delta signal}, instead of directly imitating the teacher's output distribution. The \textit{delta signal} is defined as the difference between the teacher model and its base model prior to instruction tuning for reasoning capability. 
It therefore captures the changes induced by reasoning tuning and provides a more direct signal for transferring reasoning capabilities.
Using extensive empirical evidence, we show that the \textit{delta signal} substantially improves on-policy distillation and refer to the new distillation method as On-Policy Delta Distillation (OPD$^2$). Experiments across mathematics, science, and code-reasoning benchmarks demonstrate that OPD$^2$ consistently outperforms conventional on-policy distillation, enabling reasoning LLMs to achieve strong performance with only a short post-training period. Code will be available at \url{https://github.com/naver-ai/opd2}.
\end{abstract}

\section{Introduction}

As the use of large language models (LLMs) continues to expand, the demands placed on them have become increasingly diverse, driving continuous advances in training methods.
While next-token prediction~\cite{gpt,gpt3} on vast amounts of data remains a promising paradigm for pre-training, post-training processes have been extensively explored and remain key to aligning LLMs with downstream requirements.
Post-training commonly involves supervised fine-tuning (SFT), which trains models on predefined desired outputs often generated by a larger and stronger LLM, and reinforcement learning (RL), which evaluates the quality of model-generated responses and optimizes the model toward more favorable outputs.
Recently, On-Policy Distillation (OPD) has been actively studied as an alternative to reinforcement learning. With the benefits of large, high-performing models, OPD trains the model on dense signals from model-generated responses, reducing concerns about reward design and the noisy signals RL faces.
Studies show that OPD can be competitive with RL in the post-training reasoning process, with reduced computational cost and improved performance, when a high-performing model is used as the teacher model.

On-Policy Distillation (OPD) is a variant of knowledge distillation (KD) similar to RL. Traditional KD~\cite{hinton2015kd} is designed to transfer outputs from a high-performing teacher model to a student model. It is combined with the generative nature of LLMs and splits into two sub-categories.
The first way is to distill from teacher-generated sequences. 
This approach was originally named Sequence-KD~\cite{kim2016sequencekd} and has been generalized to Supervised Fine-Tuning (SFT) on large model-generated sequences.
It is widely used for the post-training stage just before RL.
The second way is the on-policy distillation of student-generated sequences.
For on-policy training, it generates sequences using the student network and uses the teachers to score the sampled tokens.
The generated tokens that agree with the teacher network are enhanced, while tokens with low probability on the teacher network are degraded.
As in RL, the learning signal is applied only to the sampled tokens; thus, the student's original knowledge and ability are preserved throughout the post-training process.
There are extensive studies~\cite{li2025bild,peng2025pre,ma2026mopd} that extend the OPD's learning signals beyond sampled tokens, but we will not address these variants in this paper.

\begin{figure}[t]
    \centering
    \includegraphics[width=\linewidth]{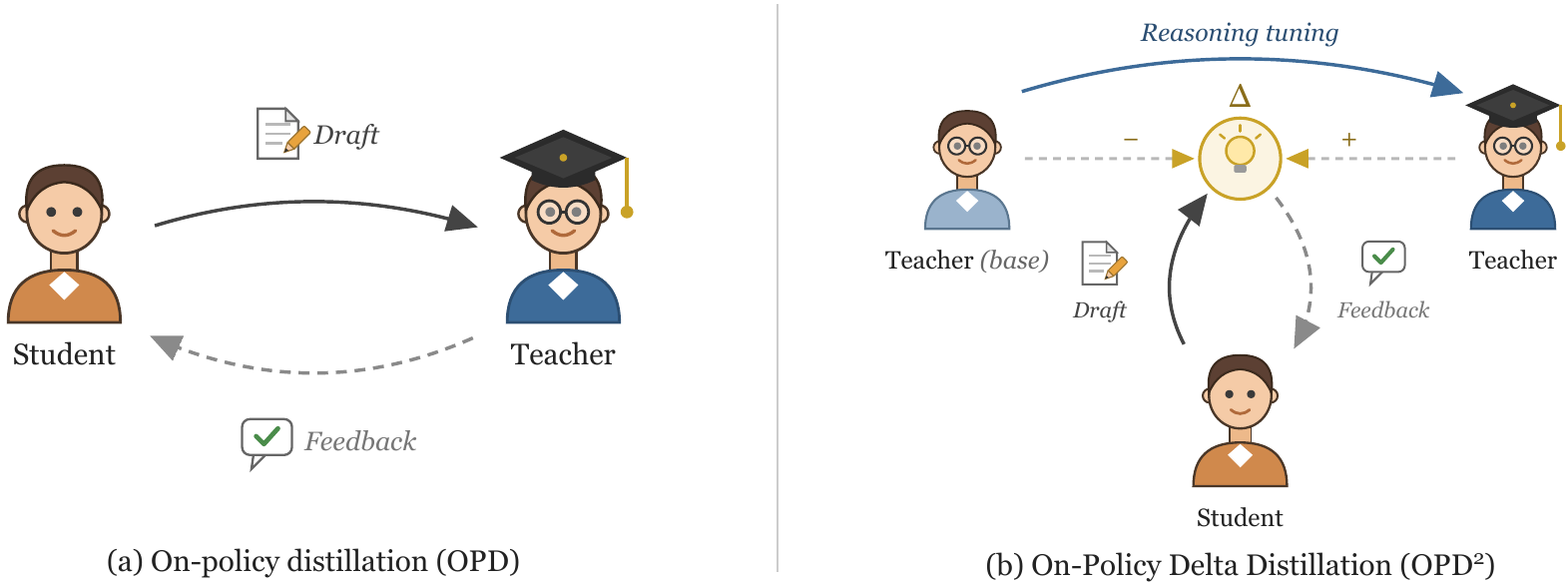}
    \caption{
        \textbf{Comparison of On-Policy Distillation (OPD) and our OPD$^2$}. While conventional on-policy distillation trains students to follow the teacher's outputs, our OPD$^2$ utilizes the \textit{delta signal}, the difference between the teacher and its base model, to focus on the knowledge of the learning trajectory required for reasoning capability.
    }
    \label{fig:concept}
\end{figure}

Despite its importance and prominence in post-training research, OPD's fundamental area remains underexplored.
The rewards for OPD are simply set to log the difference in probability between the teacher and the student.
Even though OPD achieves remarkable performance with the original loss function, there may be room for improvement in the fundamental design of its loss function.
Among the various directions for exploration, we focus on the learning trace of reasoning tuning~\cite{mitchell2023emulator}.
Most LLMs have multiple stages for training. It starts with pre-training on next-token prediction tasks using massive amounts of data. Then, tuned for reasoning and chat ability with post-training such as SFT and RL. 
As a post-training method, OPD only requires knowledge from the post-training phase. 
Thus, as shown in Fig.~\ref{fig:concept}, we design a new loss function for on-policy distillation based on the difference between the base and the reasoning tuning model of the teacher network.

Our core concept is described in Fig.~\ref{fig:concept}. In OPD, the teacher model acquires reasoning ability through reasoning tuning (\eg instruction tuning).
However, the reasoning-tuned teacher also retains natural preferences and stylistic tendencies acquired before the reasoning tuning.
We argue that distilling the reasoning-tuned teacher alone does not necessarily reveal the learning trajectory that led the teacher to acquire reasoning capabilities. 
Thus, we can provide reasoning knowledge by utilizing the delta between the reasoning-tuned teacher and the base teacher (\ie the model before post-training). 
We propose On-Policy Delta Distillation (OPD$^2$), defining the \textit{delta signal} using it as the main reward for OPD. 
In contrast to OPD, which uses the log probability difference between teacher and student as the reward signal, OPD$^2$ uses the difference between teacher and teacher base as the reward signal.
We first conduct three analyses to show the differences and benefits of the \textit{delta signal} relative to the original reward: word clouds for distillation signals, token-level visualizations, and word-level statistical analyses.
These analyses highlight the differences and characteristics of $\Delta$, providing insight into changes in reward.
Based on these analyses, we find that the \textit{delta signal} offers an effective alternative distillation signal with distinct and desirable properties.

Building on this observation,  we introduce two reward designs: centering and joint conditioning, to effectively utilize the \textit{delta signal} for OPD. First, centering is subtracting the expected rewards from the policy model's sampling probabilities to handle the signals more easily.
In the original OPD, it is automatically removed through training on-policy probabilities. But, hard to figure out the actual learning signal's direction and causes negative effects without a single sign-biased condition.
Using zero-centered rewards, we introduce a joint condition between OPD and the \textit{delta signal} to address the convergence-point issue in delta-signal-based distillation.
With these two design points, we complete the rewards design for our OPD$^2$ and successfully introduce the \textit{delta signal} into the OPD framework.

In our experiments, we develop an extensive verification framework across three reasoning domains: Math, Science, and Code. We construct a mixed-domain training set by sampling an equal number of questions from one dataset for each of the Math~\cite{openmathreasoning}, Science~\cite{opensciencereasoning2}, and Code~\cite{opencodereasoning} domains, resulting in a 1:1:1 domain ratio. We then perform on-policy distillation on this mixed training set.
The models trained by on-policy distillation methods are evaluated on 7 Math, 3 Science, and 4 Code benchmarks.
Our verification covers various sizes of Qwen3~\cite{qwen3}, both non-thinking and thinking modes, and the recently released Gemma-4~\cite{gemma4}.
In extensive verification, OPD$^2$ consistently outperforms other on-policy distillation methods, demonstrating an impressive performance gap.
It is noteworthy that OPD$^2$ significantly improves on-policy distillation across various sizes (1.7B - 8B), both in no-think and think modes, on cutting-edge models with superior reasoning performance~\cite{qwen3,gemma4}.

\section{Method}

\subsection{Preliminary}

On-Policy Distillation (OPD), a post-training method for reinforcement learning, such as GRPO, is based on efficient per-sample training costs with token-level supervision.
Qwen3 highlights the practical efficiency of OPD on its small models~\cite{qwen3}.
Subsequent studies~\cite{xiao2026mimov2,exopd} have developed detailed OPD formulations and implementations.
Basically, knowledge distillation trains the student to reduce the output probability distance from the teacher.
Since the output is a probability, Cross-Entropy or Kullback-Leibler divergence is employed for the distance function.
The loss function of OPD is defined as knowledge distillation loss on on-policy data, \ie the student's rollout samples as

\begin{equation}
\mathcal{L}_{\mathrm{OPD}}(\theta)
=
\mathbb{E}_{x \sim \mathcal{D},\, y \sim \pi_{\theta}(\cdot \mid x)}
\left[
D_{\mathrm{KL}}
\left(
\pi_{\theta}(y \mid x)
\,\middle\|\,
\pi^{*}(y \mid x)
\right)
\right],
\label{eq:pre_opd_loss}
\end{equation}

where $\pi_{\theta}$ indicates the student and $\pi^{*}$ is the teacher model. For a given question $x$, OPD generates a student response $y \sim \pi_{\theta}(\cdot \mid x)$ and trains the student's next-token prediction to mimic that of the teacher by minimizing KL divergence. 

OPD is often used to replace reinforcement learning in the post-training stage. Thus, OPD implementation also follows an RL-like framework to leverage optimized on-policy training codes and settings.
Similar to the RL setting, the OPD objective is defined as a reward applied only to sampled tokens.
For the question $x$ and given context $y_{<t}$, the reward for $t$-th token $y_t$ is defined as

\begin{equation}
R_t
=
\log \pi^{*}(y_t \mid x, y_{<t})
-
\log \pi_{\theta}(y_t \mid x, y_{<t}).
\label{eq:pre_opd_reward}
\end{equation}

Note that the reward $R_t$ is only applied to determine whether to enhance or degrade the sampled token $y_t$. It doesn't make gradients for unsampled probabilities. While there are studies on top-k OPD that address unsampled tokens for both rewards and training, we will not cover them in the paper. 
Eq.~\ref{eq:pre_opd_reward} is obtained from the partial derivative of the KL divergence with respect to the sampled-token probability $\pi_\theta(y_t \mid x, y_{<t})$. Since the reward is maximized whereas the KL divergence is minimized, we reverse the sign and omit the additive constant $-1$.

Using $R_t$ as token-level rewards, OPD trains the student with the following RL-like gradient computation:
\begin{equation}
\nabla_{\theta} J_{\mathrm{OPD}}(\theta)
=
\mathbb{E}_{x \sim \mathcal{D},\, y \sim \pi_{\theta}(\cdot \mid x)}
\left[
\sum_{t=1}^{T}
R_t
\nabla_{\theta}
\log \pi_{\theta}(y_t \mid x, y_{<t})
\right].
\label{eq:pre_opd_rl}
\end{equation}

With this implementation, the student network is trained to maximize the token-level rewards $R_t$, which is equivalent to minimizing the KL divergence between the teacher and the student on on-policy data.

\subsection{Delta Signal for Distillation}

OPD uses the log-probability difference on the sampled logit $y_t$ as rewards for the token as

\begin{equation}
R_t^\mathrm{OPD}
=
\log \pi^{*}(y_t \mid x, y_{<t})
-
\log \pi_{\theta}(y_t \mid x, y_{<t}),
\label{eq:opd_reward}
\end{equation}

where $\pi^{*}$ presents the teacher model and $\pi_{\theta}$ means the student model. 
Following the teacher's signal is a fundamental approach to knowledge distillation. But, we propose to go one step further with the base model (\ie the model prior to post-training) of the teacher $\pi^{*}_{base}$. In other words, we define the  \textit{delta signal}: 

\begin{equation}
R_t^{\Delta}
=
\log \pi^{*}(y_t \mid x, y_{<t})
-
\log \pi^{*}_{base}(y_t \mid x, y_{<t})
\label{eq:delta_reward}
\end{equation}

and use it as a primary learning objective of on-policy distillation.
$R_t^{\Delta}$ represents the learning trace of the teacher model from its own base model. In general, LLMs are pre-trained on next-token prediction on massive amounts of data, then tuned with SFT and RL to improve their reasoning ability. As shown in Fig~\ref{fig:concept}, the target knowledge for on-policy distillation is reasoning ability, not basic next-token prediction. Thus, in our on-policy distillation, $R_t^{\Delta}$ serves as the primary learning signal, extracting the teacher's knowledge to enable complex reasoning beyond simple next-token prediction. 

Before applying $R^\Delta$ for on-policy distillation, we analyze the difference between $R^{OPD}$ and $R^\Delta$ to give intuition behind this change. We provide three analyses of $R^\Delta$: word clouds to show signal patterns, signal differences with a simple question, and statistical signal changes from $R^{OPD}$ to $R^\Delta$.

\begin{figure}[t]
    \centering
    \begin{subfigure}[b]{0.325\textwidth}
        \includegraphics[width=\linewidth]{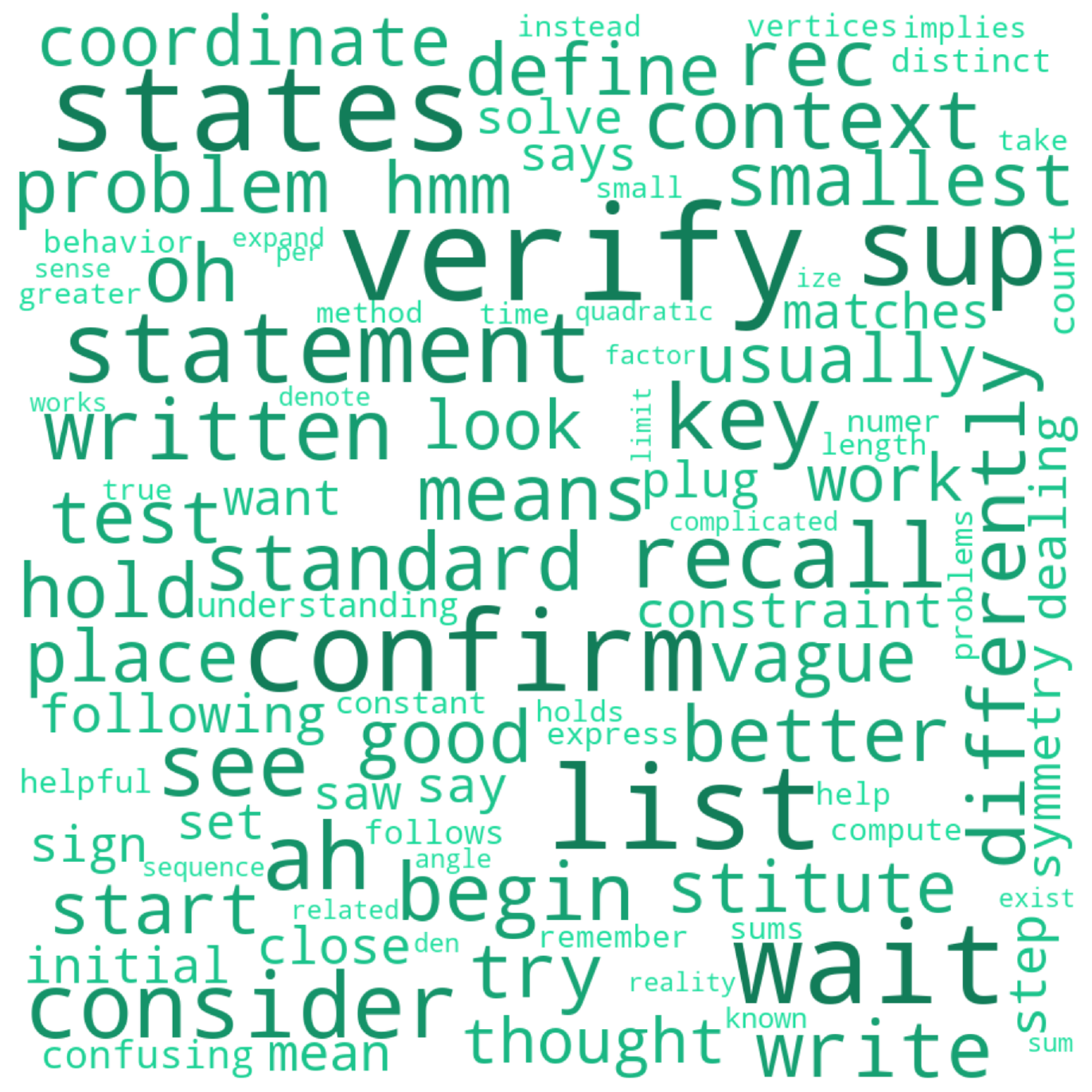}
        \caption{OPD (log$\pi^{*}$ - log$\pi_{\theta}$)}
        \label{fig:curve_aime24}
    \end{subfigure}
    \hfill
    \begin{subfigure}[b]{0.325\textwidth}
        \includegraphics[width=\linewidth]{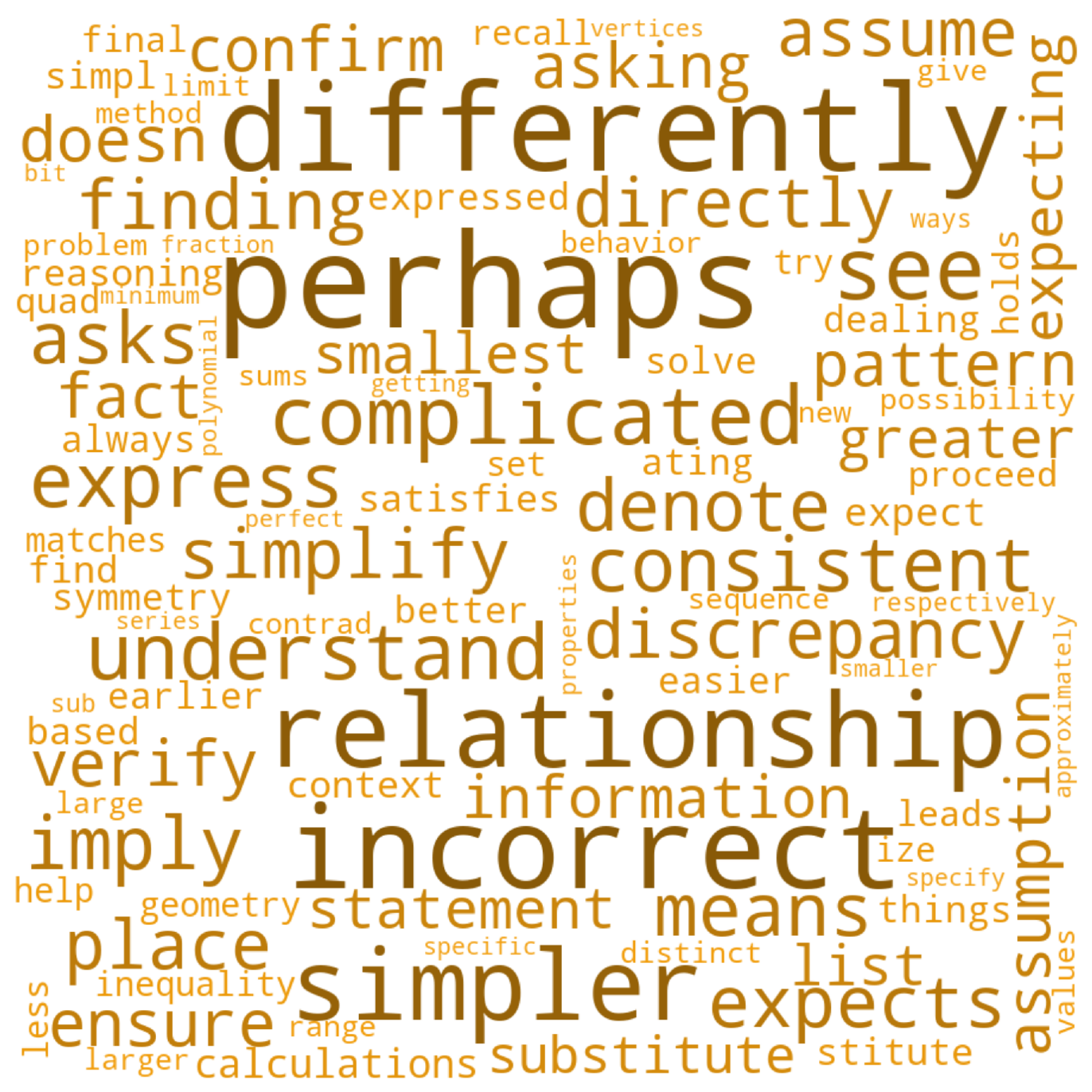}
        \caption{Base (log$\pi^{*}_{base}$ - log$\pi_{\theta}$)}
        \label{fig:curve_aime25}
    \end{subfigure}
    \hfill
    \begin{subfigure}[b]{0.325\textwidth}
        \includegraphics[width=\linewidth]{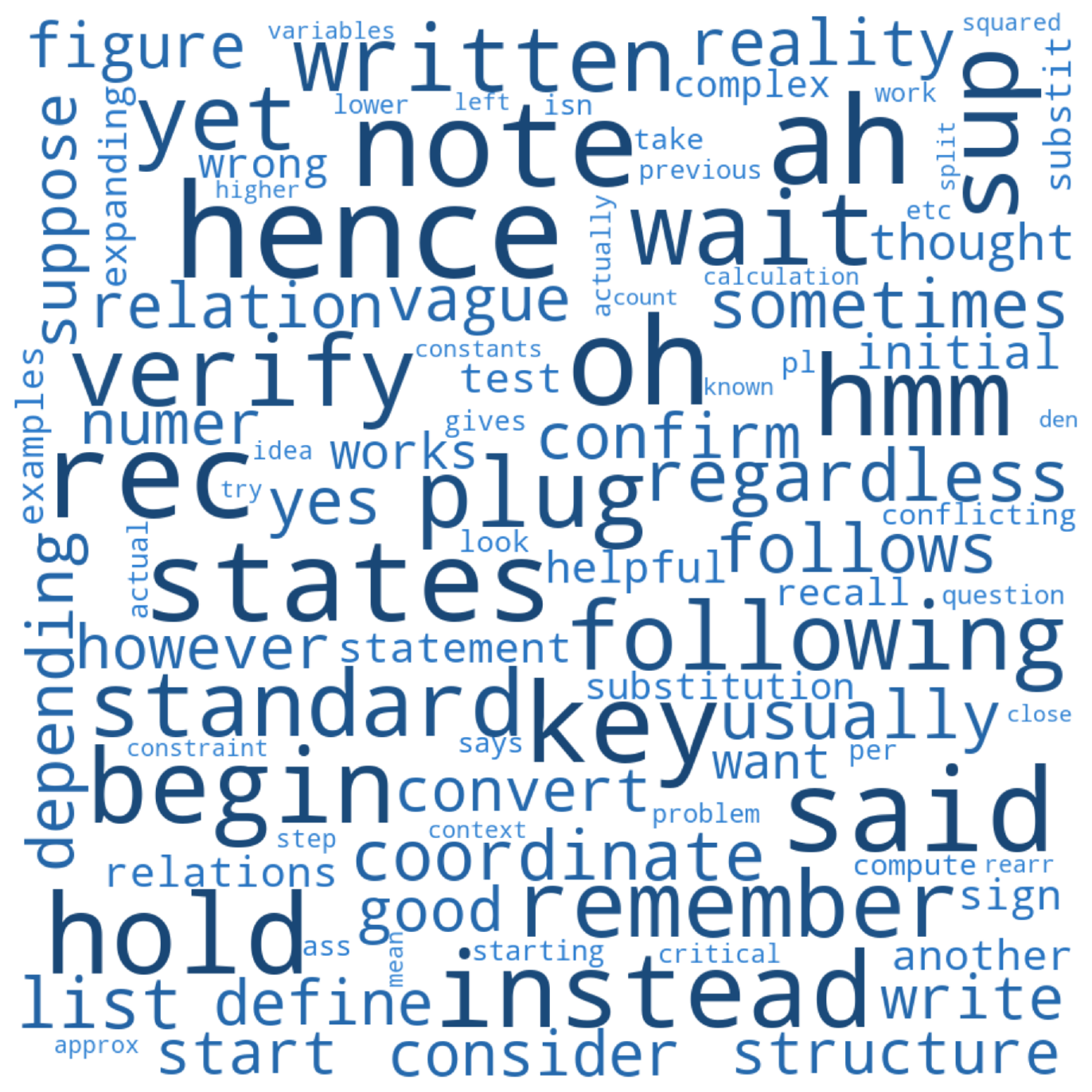}
        \caption{$\Delta$ (log$\pi^{*}$ - log$\pi^{*}_{base}$)}
        \label{fig:curve_gpqa}
    \end{subfigure}

    \vspace{0.5em}

    \caption{\textbf{Word clouds for OPD, Base and $\Delta$.} The figure illustrates distillation signal strengths in the form of word clouds. Qwen3-1.7B and Qwen3-4B are used for student and teacher, respectively, on 10k math questions. Compared with OPD, $\Delta$ emphasizes reasoning-connective words such as \textit{hence}, \textit{however}, and \textit{instead}, while suppressing exploratory and verification-related words such as \textit{see}, \textit{try}, and \textit{verify}.}
    \label{fig:cloud}
\end{figure}

\paragraph{Word clouds.} Fig.~\ref{fig:cloud} shows the representative words with strong signal patterns for teacher, base, and delta. For the analysis, we employ Qwen3-1.7B as the student and Qwen3-4B-Thinking-2507 as the teacher. Also, Qwen3-4B-Base is selected as the base model for the teacher. To simulate an OPD environment, we sample 10k problems from OpenMathReasoning~\cite{openmathreasoning} and generate responses with Qwen3-1.7B. The generated tokens are used to build OPD (log$\pi^{*}$ - log$\pi_{\theta}$), base (log$\pi^{*}_{base}$ - log$\pi_{\theta}$) and $\Delta$ (log$\pi^{*}$ - log$\pi^{*}_{base}$) signals. The signals are illustrated as word clouds using intensity on positive signals. Note that since distillation signals are highly biased to negative, we apply centering as in Eqs.~\ref{eq:opd_adv} and \ref{eq:delta_adv}.

Fig.~\ref{fig:cloud} shows the signal pattern for OPD, Base, and $\Delta$. OPD represents words enhanced by conventional on-policy distillation loss. These words are preferred by the teacher model among the sampling candidates in the student policy. Base signal means the signals preferred by the base model, which is not connected with reasoning ability and described as a negative signal in Fig.~\ref{fig:concept}. We conjecture that these natural preferences are also included in the teacher model and don't need to be transferred to the student for reasoning performance. The last $\Delta$ presents words enhanced when distillation is conducted with Eq.~\ref{eq:delta_reward}.
Compared to OPD, $\Delta$ enhances reasoning connection words such as \textit{hence}, \textit{note}, \textit{instead}, \textit{however}, and \textit{yet}. On the other hand, words that verify and explore the solution are degraded from OPD: \textit{see}, \textit{try}, \textit{verify}, and \textit{confirm}. As expected, the degraded words also appear in the base model's word cloud.

\begin{figure}[t]
    \centering
    \includegraphics[width=\linewidth]{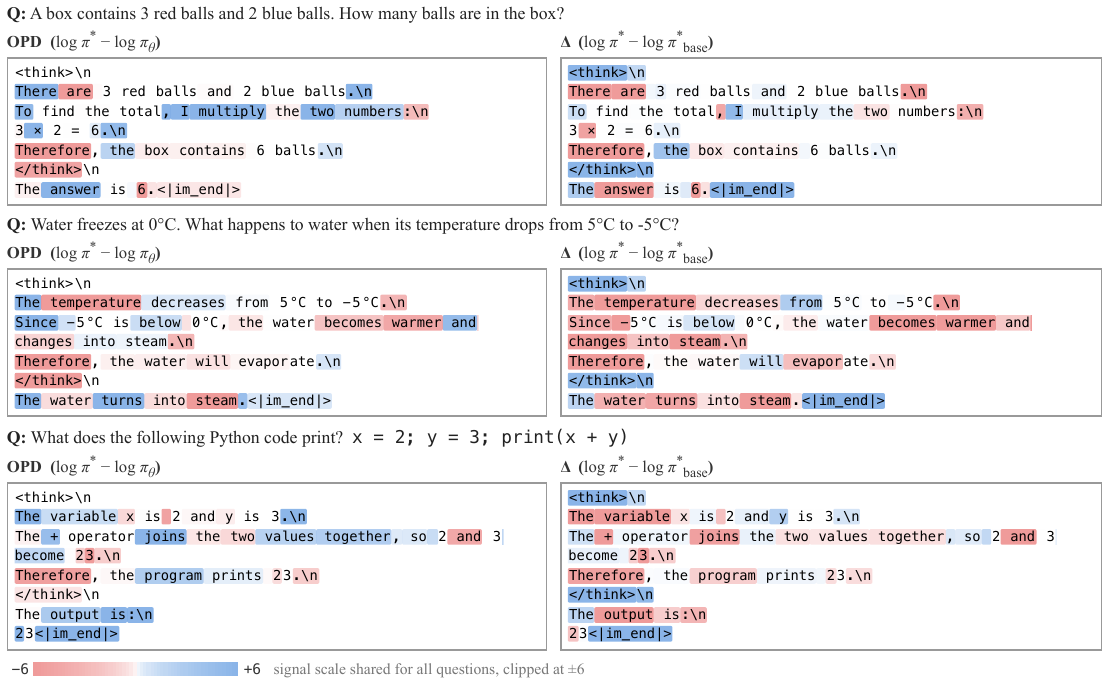}
    \caption{
    \textbf{Token-level distillation signals on simple reasoning examples.} Blue and red indicate promoting and suppressing signals, respectively, with values clipped to $\pm 6$. Compared with OPD, $\Delta$ more consistently suppresses tokens associated with incorrect reasoning.
    }
    \label{fig:token_example}
\end{figure}
\paragraph{Token-level signals.} The most representative characteristics of on-policy distillation are the token-level rewards and supervision. It is hard to achieve with other RL-based on-policy training.
We provide examples to show the token-level signal difference between OPD and $\Delta$.
It would be close to the actual training scenario that uses challenging questions.
But, the challenging scenario requires high-level knowledge and is hard to figure out the wrong parts.
Thus, we use simple reasoning questions with intentionally incorrect reasoning and answers to demonstrate how OPD and $\Delta$ signals cope with the incorrect reasoning.
The incorrect reasoning and answer were synthetically generated using an external LLM and manually curated for clarity. They are fixed inputs rather than rollouts sampled from either the student or the teacher.
The experiment is conducted for three simple questions from three domains: math, science, and code.
The question, reasoning, and answer inputs to the student (Qwen3-1.7B), teacher (Qwen3-4B-Thinking-2507), and base (Qwen3-4B-Base) to build the distillation signals $R_t^{OPD}$ and $R_t^\Delta$.
Fig.~\ref{fig:token_example} shows the token-level reward signals for OPD and $\Delta$. Blue indicates promoting signals that increase the token probability, while red indicates suppressing signals. To improve visualization, we clip the signal at $\pm$6 as shown in the color bar in Fig.~\ref{fig:token_example}.

OPD and $\Delta$ show a meaningful difference in various areas of reasoning. In general, $\Delta$ is more negative than OPD on generic reasoning expressions such as \textit{There}, \textit{Since}, and \textit{The variable}. In the first question, the reasoning goes wrong from \textit{multiply}. At those parts, $\Delta$ shows relatively negative signals compared to OPD including the "$\times$" token. It is similar to other questions. In the second, \textit{into steam} is positive on OPD while negative on $\Delta$. The critical reasoning parts in the third question are \textit{joins} and \textit{values together}, which also show different signal directions for OPD and $\Delta$. 
OPD keeps showing positive signals in the incorrect reasoning part, which is due to the magnitude difference.
The teacher and student present strong negative signals on that part. 
But, simply, student signals are stronger than those of the teacher.  
As a result, OPD produces positive signals even in traces of wrong reasoning.
In the case of $\Delta$, two models originate from the same weights, and the reasoning-tuned teacher is more sensitive to incorrect tokens.
Thus, the teacher and base model have a superior-inferior relationship in reasoning about related words, which makes $\Delta$ more robust than OPD in the correlation between the distillation signal and reasoning correctness.

\paragraph{Statistical Analysis.} 
We further analyze token-level signal changes using statistical values when the distillation signal is changed from OPD to $\Delta$.
For this analysis, Qwen3-8B is used for the student, Qwen3-30B-A3B-Thinking-2507 for the teacher, and Qwen3-30B-A3B-Base for the base.
We select a large model setting to show the clear signal statistics reducing noisy signals from failure reasoning traces.
We use the questions from OpenMathReasoning~\cite{openmathreasoning} for the math domain, OpenCodeReasoning~\cite{opencodereasoning} for the code domain, and OpenScienceReasoning-2~\cite{opensciencereasoning2} for the science domain.
In each domain, we randomly sample 10k questions and generate reasoning responses with the student model.
The generated reasonings are evaluated on the teacher and base models and converted into distillation signals: OPD and $\Delta$.
For each token in the vocabulary, we measure how often its signal changes by at least 1 when replacing OPD with $\Delta$. We separately count increases and decreases as enhanced and suppressed cases, respectively.
The thresholds of $+1$ and $-1$ correspond approximately to the top and bottom 4\% of OPD signal values and 5\% of $\Delta$ signal values, respectively. For example, an enhanced ratio of 50\% indicates that half of the occurrences of a token receive a positive signal change of at least 1. Such a token is thus strongly biased toward enhancement due to changes in the distillation signal.

Table~\ref{tab:word_stats} shows the words with high distillation strength changes when the on-policy distillation reward is replaced with $\Delta$. We conduct analysis on three domains, Math, Code, and Science, where each domain contains 10k questions, and the number of generated tokens is denoted beside the domain name.
Across domains, $\Delta$ tends to enhance explicit logical connection words: \textit{hence}, \textit{thus}, \textit{however}, and \textit{regardless}. In suppressed words, a vague expression of uncertainty, \textit{perhaps}, is commonly suppressed with a high ratio.
Also, $\Delta$ suppresses words frequently used in generic problem-solving narration, including \textit{tackle}, \textit{look}, \textit{consider}, \textit{analyze}, and \textit{computing}.
In Math, \textit{note}, \textit{why}, and \textit{depending} are enhanced, suggesting stronger emphasis on observation, verification, and conditional reasoning, while conversational expressions such as \textit{i'm} and \textit{we're} are suppressed. In Code and Science, reflective expressions such as \textit{imagine}, \textit{oh}, \textit{hmm}, and \textit{another} are also more frequent, suggesting increased consideration of alternative reasoning paths.

\begin{table}[t]
\centering
\small
\caption{
\textbf{Distillation signal difference statistics from OPD to $\Delta$.} The table shows tokens in vocabularies which is highly biased to enhanced or suppressed when the distillation signal is replaced with the \textit{delta signal}.
The ratio represents the token occurrence rate with top-5\% strength, and count denotes the total word count across all reasoning traces.
}
\label{tab:word_stats}
\begin{tabular}{
@{}r
lrr
@{\hspace{1.2em}}lrr
@{\hspace{1.2em}}lrr
@{}
}
\toprule
& \multicolumn{3}{c}{\textbf{Math} ($72.4$M)}
& \multicolumn{3}{c}{\textbf{Code} ($53.3$M)}
& \multicolumn{3}{c}{\textbf{Science} ($36.7$M)}
\\
\cmidrule(lr){2-4}
\cmidrule(lr){5-7}
\cmidrule(l){8-10}
Shift
& Word & Ratio & Count
& Word & Ratio & Count
& Word & Ratio & Count
\\
\midrule

\multirow{6}{*}{\rotatebox[origin=c]{90}{\textbf{Enhanced}}}
& \texttt{\textbf{hence}}       & 57\% & 5.0k
& \texttt{\textbf{imagine}}     & 74\% & 1.9k
& \texttt{\textbf{hence}}       & 60\% & 4.9k
\\
& \texttt{\textbf{however}}     & 55\% & 48.0k
& \texttt{\textbf{however}}     & 61\% & 28.6k
& \texttt{\textbf{hmm}}         & 49\% & 3.5k
\\
& \texttt{\textbf{note}}        & 53\% & 8.7k
& \texttt{\textbf{thus}}        & 60\% & 6.1k
& \texttt{\textbf{another}}     & 45\% & 10.8k
\\
& \texttt{\textbf{why}}         & 52\% & 2.5k
& \texttt{\textbf{oh}}          & 60\% & 2.0k
& \texttt{i\textbf{'ve}}        & 44\% & 0.8k
\\
& \texttt{\textbf{depending}}   & 46\% & 2.6k
& \texttt{1\textbf{?),}}        & 59\% & 2.1k
& \texttt{\textbf{lists}}       & 43\% & 0.9k
\\
& \texttt{\textbf{regardless}}  & 45\% & 2.9k
& \texttt{\textbf{regardless}}  & 53\% & 1.5k
& \texttt{\textbf{however}}     & 39\% & 46.0k
\\

\midrule

\multirow{6}{*}{\rotatebox[origin=c]{90}{\textbf{Suppressed}}}
& \texttt{\textbf{perhaps}}      & 49\% & 24.7k
& \texttt{\textbf{tackle}}       & 73\% & 0.8k
& \texttt{\textbf{entirely}}     & 57\% & 1.7k
\\
& \texttt{i\textbf{'m}}          & 47\% & 1.1k
& \texttt{\textbf{computing}}    & 34\% & 0.6k
& \texttt{\textbf{perhaps}}      & 38\% & 20.6k
\\
& \texttt{we\textbf{'re}}        & 41\% & 0.7k
& \texttt{\textbf{sup}pose}      & 34\% & 1.3k
& \texttt{\textbf{look}}         & 31\% & 1.2k
\\
& \texttt{\textbf{incorrect}}    & 38\% & 0.6k
& \texttt{\textbf{incorrect}}    & 32\% & 1.0k
& \texttt{\textbf{wants}}        & 30\% & 0.8k
\\
& \texttt{\textbf{see}}          & 36\% & 7.7k
& \texttt{\textbf{consider}}     & 31\% & 5.6k
& \texttt{\textbf{analyze}}      & 29\% & 0.6k
\\
& \texttt{\textbf{differently}}  & 36\% & 0.7k
& \texttt{\textbf{denote}}       & 31\% & 1.7k
& \texttt{\textbf{designed}}     & 29\% & 1.4k
\\

\bottomrule
\end{tabular}
\end{table}

\subsection{On-policy Delta Distillation (OPD$^2$)}

Although the \textit{delta signal} ($R^\Delta$) can be beneficial for distillation, it may have an issue with its convergence behavior. Specifically, unlike the original reward, $R^\Delta$ does not consider the student signal $\pi_{\theta}$ in the reward computation; as a result, the training based on $R^\Delta$ converges to the maximum reward token with a one-hot vector.
Even though this convergence point is not reachable in a practical strong-to-weak setting, it might still cause instability during training. 
Thus, we design a convergence point for $R^\Delta$ such that no gradients are computed when the student models match the teacher.

On-policy distillation shares the on-policy nature with RL. Thus, a bias invariant to the action (\ie token) in the rewards is ineffective for training~\cite{REINFORCE,sutton1999policy,mnih2016asynchronous}.  We therefore begin with the bias-subtracted OPD formulations, which obtain the advantage by subtracting the expected reward under the on-policy distribution:

\begin{equation}
A_t^\mathrm{OPD}
=
R_t^\mathrm{OPD} - \mathbb{E}_{\tilde{y}_t \sim \pi_{\theta}(\cdot \mid  x, y_{<t})} 
\left[ 
\log \pi^{*}(\tilde{y}_t \mid x, y_{<t})
-
\log \pi_{\theta}(\tilde{y}_t \mid x, y_{<t})
\right],
\label{eq:opd_adv}
\end{equation}

\begin{equation}
A_t^\Delta
=
R_t^\Delta - \mathbb{E}_{\tilde{y}_t \sim \pi_{\theta}(\cdot \mid  x, y_{<t})} 
\left[ 
\log \pi^{*}(\tilde{y}_t \mid x, y_{<t})
-
\log \pi^{*}_{base}(\tilde{y}_t \mid x, y_{<t})
\right].
\label{eq:delta_adv}
\end{equation}

$A_t^\mathrm{OPD}$ and $A_t^{\Delta}$ presents rewards advantages of sampled action $y_t$ over expected rewards values over its original probability $\pi_{\theta}(\cdot \mid  x, y_{<t})$.
During implementation, we compute the expected reward over the top-k (k=1024) tokens of $\pi_{\theta}$ to save GPU memory.
We use $A_t^\Delta$ to replace $R_t^{\mathrm{OPD}}$ for the on-policy distillation driven by $\Delta$ signals. For the convergence point issue, we add a stop condition in alignment with $A_t^\mathrm{OPD}$ as

\begin{equation}
A_t^{D^2} = 
\begin{cases}
A_t^{\Delta}
&
\text{if }
A_t^{\Delta} A_t^{\mathrm{OPD}}
>
0,
\\
0
&
\text{otherwise.}
\end{cases}
\label{eq:opd2}
\end{equation}

$A_t^{D^2}$ is our advantage function for OPD$^2$.
It basically follows $A_t^{\Delta}$ but is turned off when it conflicts with original distillation signals. The condition ($A_t^{\Delta} A_t^{\mathrm{OPD}} > 0$) prevents the student from being over-trained by the trace signal and being distant from the teacher's signal. 
When the student probability is the same as that of the teacher, \ie $\pi_\theta = \pi^{*}$, the advantage $A_t^{D^2}$ is turned off by the condition.
On-policy distillation with $A_t^{D^2}$ restricts updates to sign-consistent common descent directions between the trace $A_t^{\Delta}$ and distillation $A_t^{\mathrm{OPD}}$ signals, while using $A_t^{\Delta}$ to control the gradient magnitude. Thus, in RL form, the training gradient is described as

\begin{equation}
\nabla_{\theta} J_{\mathrm{OPD^2}}(\theta)
=
\mathbb{E}_{x \sim \mathcal{D},\, y \sim \pi_{\theta}(\cdot \mid x)}
\left[
\sum_{t=1}^{T}
A_t^{D^2}
\nabla_{\theta}
\log \pi_{\theta}(y_t \mid x, y_{<t})
\right].
\end{equation}

The effectiveness of OPD$^2$ is demonstrated in the upcoming experiment section, with extensive evaluation benchmarks across various models compared to OPD and the advanced on-policy distillation method, ExOPD.

\section{Experiment}

We design an extensive evaluation framework to verify the performance improvements of our OPD$^2$.
Since on-policy distillation is primarily used as a post-training process, we target small reasoning models, such as Qwen3~\cite{qwen3} and Gemma4~\cite{gemma4}, as student models, using large models from the same family as the teacher. As a post-training process, a notable benefit of on-policy distillation is that it can be applied to any domain without requiring domain-specific reward design. To elaborate on this advantage, we build a multi-domain training dataset by mixing sampled questions from three reasoning domains: Math, Science, and Code. After on-policy distillation, the trained model is evaluated on various benchmarks, including 7 Math, 3 Science, and 4 Code benchmarks. Note that our benchmark covers both non-thinking and thinking modes for the Qwen3 architecture. We also report training dynamics, an ablation study, and computational analysis for OPD$^2$, which help to understand detailed characteristics of OPD$^2$ training.

\subsection{Experiment settings}

\paragraph{Models.} We target open-source models with strong reasoning abilities: Qwen3~\cite{qwen3} and Gemma4~\cite{gemma4}.
Note that all student and teacher models are instruct-tuned, while the teacher-base model is -Base.
Recently, on-policy distillation studies have reported non-thinking mode performance rather than thinking-mode performance for Qwen3. 
We find that both settings offer distinct and complementary value. Qwen3 is not well-tuned for reasoning tasks in non-thinking mode. 
Thus, there is enough gap for improvement, and on-policy distillation has to make substantial changes on student models to achieve comparable performance.
On the other hand, thinking mode already has top-level reasoning ability. Thus, on-policy distillation faces the challenges of improving or altering the student model without compromising its existing performance.
Therefore, we evaluate both modes for Qwen3, whereas only the thinking mode is verified for Gemma4.

For the Qwen3 family, Qwen3-1.7B, Qwen3-4B, and Qwen3-8B are used as student models, and Gemma4-E4B-it is selected for the student in Gemma4.
The models specialized for non-thinking are used as teachers for non-thinking mode case: Qwen3-4B-Instruct-2507 for Qwen3-1.7B and Qwen3-30B-A3B-Instruct-2507 for Qwen3-4B and -8B. For thinking mode, the teacher model is switched to thinking specialized models: Qwen3-4B-Thinking-2507 and Qwen3-30B-A3B-Thinking-2507. ExOPD~\cite{exopd} and our OPD$^2$ require teacher base. Thus, we utilize the corresponding base models for these methods: Qwen3-4B-Base and Qwen3-30B-A3B-Base. For Gemma4, we use Gemma-4-31B-it as the teacher for Gemma4-E4B-it, while Gemma-4-31B serves as the teacher-base. The model setting is also summarized in Table~\ref{tab:model_configure}.

\paragraph{Training data.} We introduce a multi-domain post-training scenario for on-policy distillation. Three datasets are selected for three reasoning domains: OpenMathReasoning~\cite{openmathreasoning}, OpenScienceReasoning-2~\cite{opensciencereasoning2}, and OpenCodeReasoning~\cite{opencodereasoning}. 
These datasets contain challenging questions, which are enough to extract meaningful distillation signals with on-policy distillation.
We conduct 1:1:1 balanced sampling for each domain and make a subset with 100k questions for on-policy distillation.
Since the training uses fewer than 30k samples, the model experiences only a fraction of the available questions, corresponding to less than one training epoch. Each question is used at most once.
Note that we used only the questions from these datasets, and the reasoning traces and answers included in the datasets are discarded during sampling.

\paragraph{Training setting.} We implement on-policy distillation methods, i.e., OPD, ExOPD~\cite{exopd}, and OPD$^2$, based on TRL~\cite{vonwerra2020trl}'s GRPOTrainer with proper revision.
The revision includes enabling single completion for a question, token-level signals, and disabling group normalization.
For each question, only 1 completion is generated and forwarded to student, teacher, and teacher-base models to build corresponding distillation signals. 
We use a temperature of 0.7 for softmax after the model forward function. 
The number of training steps is set to 100, and all reported results are measured at the final training step.
The expected rewards computed in the centering operation (Eq.~\ref{eq:opd_adv} and \ref{eq:delta_adv}) are conducted on top-k (k$=1024$) tokens to reduce GPU memory cost for full-vocabulary average. The maximum completion length is set to 8k, with a temperature of 0.7 and KL regularization with a reference model disabled. The models are optimized by AdamW~\cite{adamw} using a learning rate of $5 \times 10^{-6}$, cosine lr decay, and gradient clipping at 1.0.
To prevent excessively frequent gradient clipping, we uniformly scale all rewards by a factor of 0.1 for all on-policy distillation methods.
vLLM~\cite{vllm} rollout backend integrated in TRL is used to generate responses in colocate mode for Qwen3-1.7B and in server mode on 1 node for others. Note that this setting is also summarized in Table~\ref{tab:common_configure}.

\paragraph{Evaluation.}
The models trained with the on-policy distillation methods are evaluated on various benchmarks.
We select several benchmarks that are available in Evalchemy~\cite{evalchemy} or lm-eval-harness~\cite{lm-eval-harness}.
7 Math, 3 Science, and 4 Code benchmarks are selected as the evaluation set.
For math, the models are evaluated on AIME24~\cite{AIME2024}, AIME25~\cite{AIME2025}, AMC23~\cite{AMC}, HMMT25~\cite{dekoninck2026matharena}, MATH500~\cite{hendrycks2021}, OlympiadBench~\cite{he2024olympiad}, and ReasoningGym Math~\cite{stojanovski2025reasoninggym}. For coding, we use CodeContests~\cite{codecontests}, CodeForces~\cite{guha2025openthoughts}, LiveCodeBench~\cite{jain2024livecodebench}, and ReasoningGym Algorithm~\cite{stojanovski2025reasoninggym}. GPQA~\cite{rein2024gpqa}, SuperGPQA~\cite{pteam2025supergpqascalingllmevaluation}, and SciBench~\cite{wang2024scibench} are selected for science. All benchmarks use the pass@1 metric with averages across repetitions. The repetition numbers for each benchmark are listed in Table~\ref{tab:eval_config}

\begin{table}[t]
\centering
\caption{\textbf{Qwen3 non-thinking results for Math.} Table shows non-thinking mode results for various size of Qwen3. The best performance is shown in bold, and the second best is underlined.}
\label{tab:nothink_math_results}
\small
\setlength{\tabcolsep}{5pt}
\resizebox{\textwidth}{!}{%
\begin{tabular}{lcccccccc}
\toprule
\textbf{Model} & AIME24 & AIME25 & AMC23 & HMMT25 & MATH500 & Olympiad & RGMath & \textbf{Avg} \\
\midrule
Qwen3-1.7B & 14.2 & 9.4 & 41.2 & 5.0 & 68.6 & 26.0 & 79.5 & 34.8 \\
\quad + OPD      & \underline{36.7} & \underline{23.8} & 70.8 & \textbf{17.0} & 81.0 & 37.1 & \underline{90.9} & 51.0 \\
\quad + ExOPD    & 35.4 & 23.3 & \underline{75.5} & 14.7 & \underline{81.9} & \underline{37.8} & \underline{90.9} & \underline{51.4} \\
\rowcolor{ngreenlight}
\quad + OPD$^2$  & \textbf{41.0} & \textbf{28.8} & \textbf{79.5} & \underline{15.0} & \textbf{83.9} & \textbf{40.1} & \textbf{93.9} & \textbf{54.6} \\
\midrule
Qwen3-4B & 21.9 & 18.5 & 63.8 & 12.3 & 79.8 & 33.4 & 90.6 & 45.8 \\
\quad + OPD     & 57.1 & 44.6 & 90.8 & 31.7 & 88.8 & 43.2 & 91.8 & 64.0 \\
\quad + ExOPD   & \underline{59.8} & \underline{48.1} & \underline{91.2} & \underline{37.3} & \underline{90.0} & \underline{45.2} & \underline{92.8} & \underline{66.4} \\
\rowcolor{ngreenlight}
\quad + OPD$^2$ & \textbf{68.5} & \textbf{56.5} & \textbf{94.5} & \textbf{41.0} & \textbf{91.5} & \textbf{46.3} & \textbf{93.7} & \textbf{70.3} \\
\midrule
Qwen3-8B & 28.3 & 17.9 & 66.2 & 10.7 & 80.5 & 34.2 & 90.1 & 46.9 \\
\quad + OPD     & 62.9 & 46.2 & 90.2 & \underline{33.3} & 90.0 & 43.7 & \underline{94.6} & 65.9 \\
\quad + ExOPD   & \underline{69.6} & \underline{51.7} & \underline{92.8} & 30.0 & \underline{90.3} & \underline{45.1} & \textbf{95.1} & \underline{67.8} \\
\rowcolor{ngreenlight}
\quad + OPD$^2$ & \textbf{76.2} & \textbf{59.2} & \textbf{94.8} & \textbf{38.7} & \textbf{90.9} & \textbf{46.2} & \textbf{95.1} & \textbf{71.6} \\
\bottomrule
\end{tabular}
\label{tab:qwen_math}
}
\end{table}
\begin{table}[t]
\centering
\caption{\textbf{Qwen3 non-thinking results for Code and Science.} The table shows non-thinking mode results for Qwen3. The best performance is shown in bold, and the second best is underlined.}
\label{tab:nothink_code_science_results}
\small
\setlength{\tabcolsep}{6pt}
\begin{tabular}{lccccc|cccc}
\toprule
& \multicolumn{5}{c|}{\textbf{Code}} & \multicolumn{4}{c}{\textbf{Science}} \\
\cmidrule(lr){2-6} \cmidrule(lr){7-10}
\textbf{Model}
& \begin{tabular}[c]{@{}c@{}}Code\\Contests\end{tabular}
& \begin{tabular}[c]{@{}c@{}}Code\\Forces\end{tabular}
& LCBv5
& \begin{tabular}[c]{@{}c@{}}RG\\Algo\end{tabular}
& \textbf{Avg}
& GPQA
& \begin{tabular}[c]{@{}c@{}}Super\\GPQA\end{tabular}
& \begin{tabular}[c]{@{}c@{}}Sci\\Bench\end{tabular}
& \textbf{Avg} \\
\midrule
Qwen3-1.7B & 4.8 & 10.8 & 7.0 & 19.4 & 10.5 & 39.4 & 26.1 & 26.9 & 30.8 \\
\quad + OPD & 9.9 & 16.0 & 27.9 & 30.0 & 21.0 & \underline{48.4} & 26.1 & 35.0 & \underline{36.5} \\
\quad + ExOPD & \underline{16.8} & \underline{18.6} & \underline{29.2} & \underline{33.6} & \underline{24.6} & 48.0 & \underline{26.2} & \underline{35.4} & \underline{36.5} \\
\rowcolor{ngreenlight}
\quad + OPD$^2$ & \textbf{24.6} & \textbf{21.3} & \textbf{34.1} & \textbf{37.5} & \textbf{29.4}
             & \textbf{49.8} & \textbf{27.3} & \textbf{39.2} & \textbf{38.8} \\
\midrule
Qwen3-4B & 12.9 & 18.0 & 23.9 & 33.6 & 22.1 & 42.3 & 32.6 & 45.2 & 40.0 \\
\quad + OPD & 15.2 & 27.1 & 40.7 & 42.8 & 31.4 & 54.2 & \underline{34.6} & 52.7 & 47.2 \\
\quad + ExOPD & \underline{25.9} & \textbf{28.8} & \underline{40.7} & \underline{53.5} & \underline{37.2}
             & \underline{54.8} & 33.6 & \underline{55.1} & \underline{47.8} \\
\rowcolor{ngreenlight}
\quad + OPD$^2$ & \textbf{30.7} & \underline{27.6} & \textbf{40.9} & \textbf{61.1} & \textbf{40.1}
             & \textbf{60.0} & \textbf{36.4} & \textbf{55.3} & \textbf{50.5} \\
\midrule
Qwen3-8B & 15.2 & 20.6 & 27.9 & 36.6 & 25.1 & 49.1 & 34.1 & 47.3 & 43.5 \\
\quad + OPD & 17.8 & \underline{31.1} & 42.6 & \underline{48.7} & 35.0 & 61.0 & \textbf{38.9} & 49.1 & 49.7 \\
\quad + ExOPD & \underline{30.1} & 30.4 & \textbf{47.8} & 43.9 & \underline{38.1} & \underline{61.8} & 38.5 & \underline{50.4} & \underline{50.2} \\
\rowcolor{ngreenlight}
\quad + OPD$^2$ & \textbf{34.1} & \textbf{32.6} & \underline{40.8} & \textbf{52.0} & \textbf{39.9}
             & \textbf{63.8} & \underline{38.6} & \textbf{52.3} & \textbf{51.6} \\
\bottomrule
\end{tabular}%
\label{tab:qwen_other}
\end{table}

\subsection{Reasoning performance}

Based on the extensive evaluation, we compare the performance of OPD$^2$ with the original model, OPD, and ExOPD~\cite{exopd}. Since the original models are reasoning-tuned and have strong reasoning abilities, on-policy distillation methods are first evaluated to determine whether they are beneficial to the model.
OPD means basic on-policy distillation training with Eq.~\ref{eq:pre_opd_rl}. 
Although it is the most fundamental on-policy distillation method, it still shows promising performance improvements and is applied in various practical settings~\cite{qwen3,xiao2026mimov2}.
ExOPD~\cite{exopd} is a recently proposed advanced method for on-policy distillation. It leverages a teacher–base model to amplify the extrapolation signal and represents a state-of-the-art approach in this area. We implement ExOPD with $\lambda=1.25$, following the setting reported in the original paper.

\paragraph{Non-thinking mode.}
We first evaluate all methods in the non-thinking mode, where the models generate answers without explicit reasoning traces. As shown in Tables~\ref{tab:qwen_math} and~\ref{tab:qwen_other}, all on-policy distillation methods substantially improve the original Qwen3 models across Math, Code, and Science, suggesting that on-policy distillation is particularly effective at strengthening the relatively weak non-thinking capabilities of reasoning-tuned models.

Among the methods compared, OPD$^2$ achieves the best average performance across all model sizes and domains. The improvements are particularly notable in Math. For Qwen3-1.7B, OPD$^2$ increases the average score from 34.8 to 54.6, outperforming OPD and ExOPD by 3.6 and 3.2 points, respectively. Similar gains are observed for Qwen3-4B and Qwen3-8B, where OPD$^2$ achieves average scores of 70.3 and 71.6, compared with 66.4 and 67.8 for ExOPD. Notably, the 4B model trained with OPD$^2$ already surpasses the 8B model trained with both OPD and ExOPD, demonstrating that the proposed method can provide gains comparable to, or greater than, those obtained by increasing the model size.

OPD$^2$ also consistently improves performance in Code and Science. On Code, it achieves average scores of 29.4, 40.1, and 39.9 for the 1.7B, 4B, and 8B models, respectively. The gain is especially large for Qwen3-1.7B, improving the original model by 18.9 points and ExOPD by 4.8 points. On Science, OPD$^2$ obtains the highest average score for all three model sizes, reaching 38.8, 50.5, and 51.6. These results indicate that the advantages of OPD$^2$ are not restricted to mathematical reasoning but generalize across diverse reasoning domains.

\begin{table}[t]
\centering
\caption{\textbf{Qwen3 thinking results for Math.} The table shows thinking mode results for various size of Qwen3. The best performance is shown in bold, and the second best is underlined.}
\label{tab:think_math_results}
\small
\setlength{\tabcolsep}{5pt}
\resizebox{\textwidth}{!}{%
\begin{tabular}{lcccccccc}
\toprule
\textbf{Model} & AIME24 & AIME25 & AMC23 & HMMT25 & MATH500 & Olympiad & RGMath & \textbf{Avg} \\
\midrule
Qwen3-1.7B & \underline{50.4} & 32.9 & \underline{85.2} & \underline{22.3} & \underline{86.6} & 39.9 & 96.8 & \underline{59.2} \\
\quad + OPD      & 41.5 & 34.2 & 80.2 & 20.7 & 85.6 & 39.9 & \underline{97.3} & 57.1 \\
\quad + ExOPD    & 44.6 & \underline{35.4} & 84.5 & 20.3 & \underline{86.6} & \underline{40.6} & 96.9 & 58.4 \\
\rowcolor{ngreenlight}
\quad + OPD$^2$  & \textbf{51.9} & \textbf{42.7} & \textbf{89.8} & \textbf{26.3} & \textbf{88.1} & \textbf{42.1} & \textbf{97.7} & \textbf{62.7} \\
\midrule
Qwen3-4B & \textbf{73.3} & \underline{64.8} & 96.2 & \underline{44.3} & 90.9 & 45.4 & \underline{98.1} & 73.3 \\
\quad + OPD      & 65.4 & 56.0 & \underline{98.0} & 41.7 & \textbf{91.9} & 45.8 & 97.6 & 70.9 \\
\quad + ExOPD    & 68.1 & 60.2 & \underline{98.0} & 43.7 & \underline{91.8} & \underline{46.0} & \textbf{98.5} & \underline{72.3} \\
\rowcolor{ngreenlight}
\quad + OPD$^2$  & \textbf{73.3} & \textbf{66.9} & \textbf{98.2} & \textbf{48.7} & 91.5 & \textbf{47.3} & 97.7 & \textbf{74.8} \\
\midrule
Qwen3-8B & \underline{76.0} & 64.0 & 95.8 & 44.3 & 91.0 & 46.3 & 98.4 & \underline{73.7} \\
\quad + OPD      & 69.2 & 59.4 & 97.2 & 43.3 & 91.5 & 46.9 & 98.2 & 72.2 \\
\quad + ExOPD    & 71.7 & \underline{62.1} & \underline{97.5} & \underline{46.3} & \textbf{91.9} & \underline{47.1} & \textbf{98.7} & 73.6 \\
\rowcolor{ngreenlight}
\quad + OPD$^2$  & \textbf{76.5} & \textbf{66.9} & \textbf{98.8} & \textbf{52.3} & \textbf{91.9} & \textbf{47.5} & 97.7 & \textbf{75.9} \\
\bottomrule
\end{tabular}%
\label{tab:qwen_think_math}
}
\end{table}
\begin{table}[t]
\centering
\caption{\textbf{Qwen3 thinking results for Code and Science.} The table shows thinking mode results for Qwen3. The best performance is shown in bold, and the second best is underlined.}
\label{tab:think_code_science_results}
\small
\setlength{\tabcolsep}{6pt}
\begin{tabular}{lccccc|cccc}
\toprule
& \multicolumn{5}{c|}{\textbf{Code}} & \multicolumn{4}{c}{\textbf{Science}} \\
\cmidrule(lr){2-6} \cmidrule(lr){7-10}
\textbf{Model}
& \begin{tabular}[c]{@{}c@{}}Code\\Contests\end{tabular}
& \begin{tabular}[c]{@{}c@{}}Code\\Forces\end{tabular}
& LCBv5
& \begin{tabular}[c]{@{}c@{}}RG\\Algo\end{tabular}
& Avg
& GPQA
& \begin{tabular}[c]{@{}c@{}}Super\\GPQA\end{tabular}
& \begin{tabular}[c]{@{}c@{}}Sci\\Bench\end{tabular}
& Avg \\
\midrule
Qwen3-1.7B & 19.8 & 21.6 & 31.6 & 44.3 & 29.3 & 48.0 & 28.1 & \textbf{44.1} & 40.1 \\
\quad + OPD      & 25.9 & 23.9 & 34.7 & \underline{45.2} & 32.4 & 54.2 & 29.4 & 41.6 & 41.7 \\
\quad + ExOPD    & \underline{36.0} & \underline{28.0} & \underline{39.8} & 44.6 & \underline{37.1} & \underline{54.3} & \underline{29.6} & 42.6 & \underline{42.2} \\
\rowcolor{ngreenlight}
\quad + OPD$^2$  & \textbf{37.8} & \textbf{30.0} & \textbf{42.2} & \textbf{51.5} & \textbf{40.4}
               & \textbf{56.8} & \textbf{30.2} & \underline{43.4} & \textbf{43.4} \\
\midrule
Qwen3-4B & 34.8 & \textbf{40.0} & \textbf{55.9} & 63.9 & 48.7
        & \textbf{58.2} & 35.6 & \textbf{60.1} & \underline{51.3} \\
\quad + OPD   & 38.6 & 34.8 & 48.2 & 64.1 & 46.4
        & 51.4 & 35.6 & 58.1 & 48.4 \\
\quad + ExOPD & \underline{42.6} & \underline{36.3} & \underline{50.1} & \underline{67.8} & \underline{49.2}
        & 52.9 & \underline{36.0} & 58.1 & 49.0 \\
\rowcolor{ngreenlight}
\quad + OPD$^2$ & \textbf{45.3} & 36.1 & 48.4 & \textbf{73.4} & \textbf{50.8}
        & \underline{57.5} & \textbf{38.1} & \underline{58.8} & \textbf{51.5} \\
\midrule
Qwen3-8B & 39.2 & \underline{43.4} & 48.4 & \underline{72.3} & 50.8
        & \underline{64.1} & 37.2 & 57.8 & \underline{53.0} \\
\quad + OPD   & 41.0 & 41.3 & 53.8 & 66.5 & 50.7
        & 59.3 & 37.2 & 56.3 & 50.9 \\
\quad + ExOPD & \underline{46.5} & 43.6 & \underline{56.2} & 69.8 & \underline{54.0}
        & 61.1 & \underline{38.3} & 56.5 & 52.0 \\
\rowcolor{ngreenlight}
\quad + OPD$^2$ & \textbf{48.5} & \textbf{47.9} & \textbf{61.0} & \textbf{73.7} & \textbf{57.8}
        & \textbf{64.4} & \textbf{40.6} & \textbf{58.7} & \textbf{54.6} \\
\bottomrule
\end{tabular}%
\label{tab:qwen_think_other}
\end{table}

\paragraph{Thinking mode.}
We next evaluate the methods in the thinking mode, where the original Qwen3 models already exhibit strong reasoning performance. As shown in Tables~\ref{tab:qwen_think_math} and~\ref{tab:qwen_think_other}, improving these strong baselines through on-policy distillation is substantially more challenging than improving their relatively weak non-thinking capabilities. In particular, standard OPD frequently degrades performance, while ExOPD provides only limited or inconsistent improvements. For example, on Math, both methods underperform the original models for all three model sizes in terms of average performance.

In contrast, OPD$^2$ consistently achieves the highest average performance across all model sizes and domains. On Math, OPD$^2$ improves the original 1.7B, 4B, and 8B models by 3.5, 1.5, and 2.2 points, respectively, reaching average scores of 62.7, 74.8, and 75.9. The improvements are particularly notable on challenging competition-level benchmarks such as AIME25 and HMMT25. For instance, OPD$^2$ improves the HMMT25 score of Qwen3-8B from 44.3 to 52.3, whereas OPD and ExOPD achieve 43.3 and 46.3, respectively.

A similar trend is observed in Code and Science. On Code, OPD$^2$ improves the average scores of the 1.7B, 4B, and 8B models from 29.3, 48.7, and 50.8 to 40.4, 50.8, and 57.8, respectively. In particular, the 8B model obtains consistent gains across all four code benchmarks. On Science, where the original models are already competitive, OPD$^2$ still achieves the best average performance for every model size, whereas OPD and ExOPD often reduce the baseline performance. These results demonstrate that OPD$^2$ can improve not only relatively weak capabilities but also the already strong reasoning capabilities of reasoning-tuned models without sacrificing their overall performance.

\begin{table}[t]
\centering
\caption{\textbf{Gemma4 for Math.} The table shows on-policy distillation results for Gemma4-E4B-it. The best performance is shown in bold, and the second best is underlined.}
\label{tab:gemma4_think_math_results}
\small
\setlength{\tabcolsep}{5pt}
\resizebox{\textwidth}{!}{%
\begin{tabular}{lcccccccc}
\toprule
\textbf{Model} & AIME24 & AIME25 & AMC23 & HMMT25 & MATH500 & Olympiad & RGMath & \textbf{Avg} \\
\midrule
Gemma4-E4B-it
    & 51.7
    & 37.9
    & 88.8
    & 27.3
    & 87.2
    & 41.6
    & 89.8
    & 60.6 \\
\quad + OPD
    & 49.4
    & 35.6
    & 84.5
    & 25.0
    & 82.9
    & 41.5
    & 93.4
    & 58.9 \\
\quad + ExOPD
    & \underline{59.6}
    & \underline{41.7}
    & \underline{90.0}
    & \underline{38.0}
    & \underline{87.8}
    & \textbf{46.0}
    & \underline{93.9}
    & \underline{65.3} \\
\rowcolor{ngreenlight}
\quad + OPD$^2$
    & \textbf{69.2}
    & \textbf{44.0}
    & \textbf{92.3}
    & \textbf{40.0}
    & \textbf{88.3}
    & \underline{45.5}
    & \textbf{95.7}
    & \textbf{67.8} \\
\bottomrule
\end{tabular}%
\label{tab:gemma_math}
}
\end{table}
\begin{table}[t]
\centering
\caption{\textbf{Gemma4 for Code and Science.} The table shows on-policy distillation results for Gemma4-E4B-it. The best performance is shown in bold, and the second best is underlined.}
\label{tab:gemma4_think_code_science_results}
\small
\setlength{\tabcolsep}{6pt}
\begin{tabular}{lccccc|cccc}
\toprule
& \multicolumn{5}{c|}{\textbf{Code}}
& \multicolumn{4}{c}{\textbf{Science}} \\
\cmidrule(lr){2-6}
\cmidrule(lr){7-10}
\textbf{Model}
& \begin{tabular}[c]{@{}c@{}}Code\\Contests\end{tabular}
& \begin{tabular}[c]{@{}c@{}}Code\\Forces\end{tabular}
& LCBv5
& \begin{tabular}[c]{@{}c@{}}RG\\Algo\end{tabular}
& Avg
& GPQA
& \begin{tabular}[c]{@{}c@{}}Super\\GPQA\end{tabular}
& \begin{tabular}[c]{@{}c@{}}Sci\\Bench\end{tabular}
& Avg \\
\midrule
Gemma4-E4B-it
    & \textbf{52.1}
    & \textbf{43.5}
    & \textbf{63.0}
    & 62.2
    & \textbf{55.2}
    & \textbf{59.9}
    & 34.4
    & 46.8
    & 47.0 \\
\quad + OPD
    & 12.3
    & 27.4
    & 47.5
    & 60.4
    & 36.9
    & 46.0
    & 32.9
    & 38.1
    & 39.0 \\
\quad + ExOPD
    & 29.5
    & 34.1
    & \underline{53.1}
    & \underline{63.7}
    & 45.1
    & 55.4
    & \underline{36.4}
    & \underline{50.2}
    & \underline{47.3} \\
\rowcolor{ngreenlight}
\quad + OPD$^2$
    & \underline{44.4}
    & \underline{37.2}
    & 52.3
    & \textbf{64.0}
    & \underline{49.5}
    & \underline{57.8}
    & \textbf{37.7}
    & \textbf{50.9}
    & \textbf{48.8} \\
\bottomrule
\end{tabular}%
\label{tab:gemma_other}
\end{table}

\paragraph{Gemma4.}
Finally, we evaluate the methods on Gemma4-E4B-it to examine whether their effectiveness generalizes beyond the Qwen3 model family. As shown in Tables~\ref{tab:gemma_math} and~\ref{tab:gemma_other}, standard OPD substantially degrades performance across all three domains, indicating that directly applying conventional on-policy distillation can be unstable when applied to a strong model from a different family. ExOPD alleviates this degradation and yields meaningful improvements in Math, but its effectiveness remains inconsistent across domains.

OPD$^2$ achieves the best Math performance by a clear margin, improving the average score from 60.6 to 67.8. It outperforms ExOPD by 2.5 points on average and achieves the highest score on six of the seven benchmarks. The gain is particularly remarkable on AIME24, where OPD$^2$ improves the original model from 51.7 to 69.2, compared with 59.6 for ExOPD. These results show that the benefit of OPD$^2$ is not specific to Qwen3 and transfers effectively to a distinct model architecture and training recipe.

The results on Code present a more challenging case. Gemma4-E4B-it already exhibits strong coding performance, and none of the distillation methods surpasses the original model in terms of average score. Nevertheless, OPD$^2$ retains substantially more of the original capability than OPD and ExOPD, achieving an average score of 49.5 compared with 36.9 and 45.1, respectively. It also improves RGAlgo from 62.2 to 64.0. On Science, OPD$^2$ improves the average score from 47.0 to 48.8 and achieves the best results on SuperGPQA and SciBench. Overall, these results demonstrate that OPD$^2$ generalizes across model families and provides a more robust trade-off between improving target capabilities and preserving the strong capabilities of the original model.

\subsection{Training dynamics}
\begin{figure}[t]
    \centering
    \begin{subfigure}[b]{0.325\textwidth}
        \includegraphics[width=\linewidth]{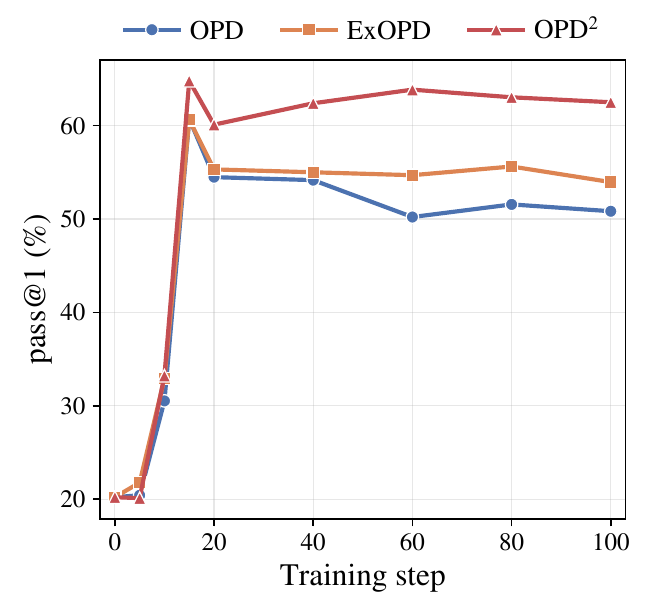}
        \caption{AIME24 \& 25}
        \label{fig:curve_aime24}
    \end{subfigure}
    \hfill
    \begin{subfigure}[b]{0.325\textwidth}
        \includegraphics[width=\linewidth]{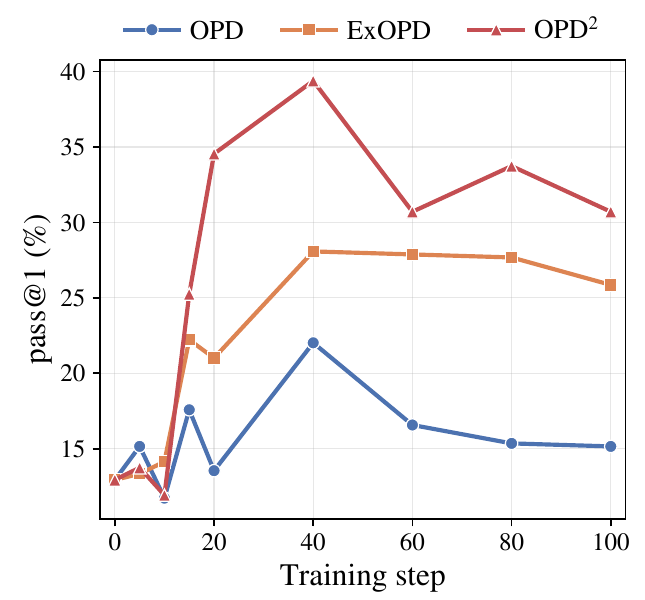}
        \caption{CodeContests}
        \label{fig:curve_aime25}
    \end{subfigure}
    \hfill
    \begin{subfigure}[b]{0.325\textwidth}
        \includegraphics[width=\linewidth]{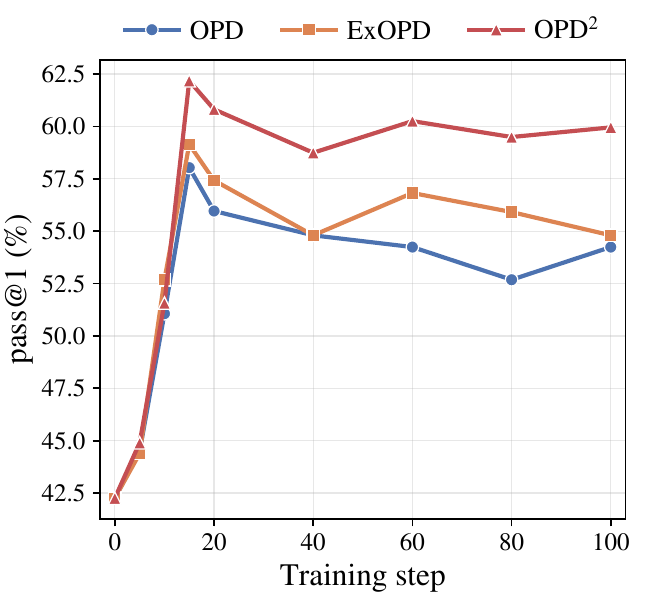}
        \caption{GPQA}
        \label{fig:curve_gpqa}
    \end{subfigure}

    \vspace{0.5em}

    \caption{\textbf{Training curve for various domains.} The figures show the reasoning performance changes during on-policy distillation training. We report the average performance on AIME24 and AIME25 for Math, CodeContests for Code, and GPQA for Science. All OPD variants exhibit similar training dynamics, while OPD$^2$ achieves the strongest performance. Note that, in all tables, we report performance at the final training step, not the peak performance observed during training.}
    \label{fig:training_step_accuracy}
\end{figure}

OPD$^2$ demonstrates substantial performance improvements across all domains and consistently outperforms the other on-policy distillation methods by a meaningful margin.
To understand how these gains emerge during training, we compare the performance trajectories of OPD, ExOPD, and OPD$^2$ under the Qwen3-4B non-thinking setting in Fig.~\ref{fig:training_step_accuracy}.
All methods exhibit a rapid increase in performance during the early stages of training, indicating that most of the benefits of on-policy distillation are achieved within a relatively small number of optimization steps.
After this initial improvement, however, their trajectories diverge considerably.

OPD and ExOPD tend to reach their peak performance early and subsequently plateau or degrade as training proceeds.
This behavior is particularly evident on Math and Code, where the final performance can be noticeably lower than the intermediate peak.
In contrast, OPD$^2$ achieves a larger initial improvement and maintains a consistently higher performance throughout the remaining training steps.
Although its performance also fluctuates after the early peak, it remains clearly above that of OPD and ExOPD across all three domains.

The separation is especially notable on CodeContests, where OPD$^2$ maintains a large advantage over the competing methods throughout training.
A similar gap is observed in Math and Science.
These results suggest that the improvements of OPD$^2$ do not arise from an isolated peak at a particular checkpoint, but reflect a persistent advantage over the entire training trajectory.
Please note that all results in the main tables are reported at the final training step. We do not select the best-performing checkpoint.

\subsection{Ablation study}
\begin{table}[t]
\centering
\small
\setlength{\tabcolsep}{6pt}
\caption{\textbf{Ablation study for OPD$^2$} We conduct an ablation study for OPD$^2$. Replacing the \textit{delta signal} shows the most significant impact. The formulas indicate how each variant modifies Eq.~\ref{eq:opd2}.}
\label{tab:opd2_ablation}
\begin{tabular}{l ccc c ccc}
\toprule
& \multicolumn{3}{c}{\textbf{Non-thinking}} & & \multicolumn{3}{c}{\textbf{Thinking}} \\
\cmidrule{2-4}\cmidrule{6-8}
 & Math & Code & Science & & Math & Code & Science \\
\midrule
OPD$^2$ & 54.6 & 29.4 & 38.8 & & 62.7 & 40.4 & 43.5 \\
\quad No \textit{delta signal} ($A_t^{\Delta} \rightarrow A_t^{\mathrm{OPD}}$)
    & 50.5 & 22.5 & 35.9 & & 57.4 & 32.1 & 41.3 \\
\quad No condition (\cancel{$A_t^{\Delta} A_t^{\mathrm{OPD}} > 0$})
    & 55.8 & 28.8 & 38.8 & & 61.5 & 39.3 & 43.8 \\
\quad No centering ($A_t^{\Delta} \rightarrow R_t^{\Delta}$)
    & 54.9 & 28.5 & 38.8 & & 63.1 & 39.3 & 43.3 \\
\bottomrule
\end{tabular}
\end{table}
We conduct an ablation study to analyze the contribution of each component in OPD$^2$. As shown in Table~\ref{tab:opd2_ablation}, replacing the \textit{delta signal} with the standard OPD signal causes the largest performance degradation in both non-thinking and thinking modes. This result indicates that the \textit{delta signal} is the primary source of the performance gains achieved by OPD$^2$.
Removing the agreement condition or the centering operation has a comparatively smaller effect. Although these components generally contribute to the overall performance, their impact is not as substantial or consistent as that of the \textit{delta signal}. Overall, the ablation results show that the \textit{delta signal} plays the central role in OPD$^2$, while the condition and centering provide additional but relatively modest improvements.

\subsection{Computational costs}
\newcommand{\tinc}[1]{{\scriptsize\,(+#1\%)}}

\begin{table}[t]
\centering
\small
\setlength{\tabcolsep}{7pt}
\caption{\textbf{Training time for on-policy distillation.} The table shows wall-clock training time (hours) spent on distillation. Parentheses show the relative increase over OPD. The time is measured on NVIDIA H100 8-GPU nodes. $^{\dagger}$Qwen3-1.7B runs on a single node, while all others run on 4 nodes.}
\label{tab:train_time_think}
\begin{tabular}{lcccc}
\toprule
\textbf{Method} & \textbf{Qwen3-1.7B}$^{\dagger}$ & \textbf{Qwen3-4B} & \textbf{Qwen3-8B} & \textbf{Gemma4-E4B} \\
\midrule
OPD      & 4.4 \tinc{0} & 7.3 \tinc{0} & 7.6 \tinc{0} & 12.7 \tinc{0} \\
ExOPD    & 5.3 \tinc{19} & 9.1 \tinc{26} & 9.4 \tinc{24} & 13.8 \tinc{9} \\
OPD$^2$  & 5.5 \tinc{24} & 9.3 \tinc{28} & 9.6 \tinc{27} & 13.8 \tinc{8} \\
\bottomrule
\end{tabular}
\end{table}

OPD$^2$ introduces additional computation compared with standard OPD because it requires a teacher-base model forward pass to construct the \textit{delta signal}. As shown in Table~\ref{tab:train_time_think}, this increases the wall-clock training time by approximately 24-28\% for the Qwen3 models and by 8\% for Gemma4-E4B. 
For OPD$^2$, this additional computation represents a modest overhead compared with standard OPD.
However, the computation cost of OPD$^2$ remains comparable to that of ExOPD, with only a small difference in training time across all model sizes. 
Our current implementation is not optimized for the reward computation in Eq.~\ref{eq:delta_reward}, suggesting that the reported overhead can be further reduced.
Moreover, on-policy distillation typically converges within a relatively small number of training steps, and prolonged training can even lead to performance degradation, as observed in Fig.~\ref{fig:training_step_accuracy}.
Thus, although OPD$^2$ requires additional computational cost compared to standard OPD, its additional overhead remains limited in the typical short-training regime of on-policy distillation.
\section{Related Work}

Post-training commonly relies on SFT and RL to align LLMs with reasoning tasks. On-policy distillation (OPD) has recently emerged as an alternative post-training method built on knowledge distillation. The distillation transfers the outputs of a high-performing teacher to a student~\cite{hinton2015distilling}. In fact, the SFT above is itself a form of KD. Its first form learns from teacher-generated sequences, named Sequence-KD~\cite{kim2016sequence}, and later generalized to SFT on large model-generated sequences~\cite{s1, openthought}. However, training on the teacher's sequences has a limitation. The student learns from the teacher's distribution but generates from its own distribution at inference. This train-inference mismatch is known as exposure bias~\cite{bengio2015scheduled}. On-policy distillation addresses this issue. Instead of teacher sequences, it uses the student's own rollout samples and lets the teacher score the sampled tokens~\cite{agarwal2024policy, gu2026minillm, ko2024distillm}. Thus, the learning signal is applied to the states the student actually visits at inference. Recent studies show that OPD is competitive with RL in reasoning post-training, with dense token-level signals and reduced cost~\cite{lu2025onpolicydistillation}, and its efficiency has been reported on small models in recent releases~\cite{qwen3,xiao2026mimov2}.

Following OPD, various studies extend or analyze the on-policy distillation framework. Some studies modify the learning signal, such as reweighting it by uncertainty~\cite{jin2026entropy} or allowing a controlled degree of off-policy data for efficiency~\cite{ko2026scaling}. Other studies remove the external teacher and use a single LLM as both the teacher and the student, in which the teacher gets a verified solution as privileged information while the student sees only the problem~\cite{zhao2026opsd, sang2026crisp}. There are also studies that analyze OPD rather than extend it~\cite{kim2026does, li2026rethinking, song2026survey}. These methods share a common point. They all set the teacher's output as the target for the student. The most related work to ours uses the teacher's base model for the distillation signal. ExOPD~\cite{exopd} uses the teacher-base model to build an extrapolation signal. It amplifies the difference between the teacher and its base with a factor $\lambda > 1$, so the student is trained beyond the teacher. Our OPD$^2$ also uses the teacher base model, but in a different way. We do not extrapolate the teacher's output. Instead, we use the difference between the teacher and its base, the \textit{delta signal}, as the main reward. The \textit{delta signal} represents the knowledge that the teacher learned from reasoning tuning. With centering and the joint condition, OPD$^2$ focuses the signal on reasoning-related tokens while keeping the student's original ability.

\section{Conclusion}

In this paper, we have examined the potential of the \textit{delta signal}, the difference between the teacher and the base model, as the major training signal for on-policy distillation. We conjectured that the \textit{delta signal} can be an effective way to transfer reasoning knowledge obtained through reasoning tuning of the teacher model, starting from base models. Through various analyses, we showed that the \textit{delta signal} includes meaningful reasoning signals and has beneficial points compared to the original distillation rewards. Based on this observation, we have proposed OPD$^2$, on-policy distillation based on the \textit{delta signal}. Our OPD$^2$ is verified on an extensive framework comprising three reasoning domains (Math, Science, and Code), 14 evaluation benchmarks, and networks of various sizes (1.7B-8B) across cutting-edge models (Qwen3 and Gemma4). In all settings, OPD$^2$ achieved substantial performance improvements over conventional on-policy distillation, demonstrating the effectiveness of the \textit{delta signal} for on-policy distillation. We believe that OPD$^2$ can serve as an important milestone in the development of on-policy distillation.

{\small
\bibliographystyle{unsrt}
\bibliography{references}}

\newpage
\appendix

\section{Appendix}

Tables~\ref{tab:model_configure} and~\ref{tab:common_configure} summarize the model-specific configurations and common optimization settings used in our experiments. We use NVIDIA H100 GPUs for all experiments. Qwen3-1.7B is trained on a single 8-GPU node, while Qwen3-4B, Qwen3-8B, and Gemma4-E4B are trained using four nodes. For the multi-node settings, 24 GPUs are used for student training and 8 GPUs are allocated to the vLLM rollout server.

For Qwen3, we use a larger model from the same model family as the teacher. The Qwen3-1.7B student uses Qwen3-4B as its teacher, while the Qwen3-4B and Qwen3-8B students use Qwen3-30B-A3B. We select the corresponding Instruct or Thinking checkpoint according to the target generation mode. Gemma4-E4B-it is evaluated only in the thinking setting and uses Gemma4-31B-it as the teacher. The corresponding pretrained checkpoints without instruction tuning are used as the teacher-base models for ExOPD and OPD$^2$.

All methods are trained for 100 optimization steps with an effective global batch size of 256 or 288. We use AdamW with a learning rate of $5\times10^{-6}$, cosine lr decay, a warmup ratio of 0.1, and a minimum learning-rate ratio of 0.1. The gradient norm is clipped to 1.0, and all the rewards are scaled by 0.1 to prevent excessive gradient clipping. We set the maximum generation length to 8192 and the sampling temperature to 0.7. The KL coefficient is set to $\beta=0$, as we do not apply an additional KL regularization term with the reference model during training.

\begin{table}[h]
\centering
\caption{\textbf{Model configuration \& model-specific setting.} The table shows teacher and student model configuration with model-specific parameters. We use NVIDIA H100 GPU for all experiments.}
\label{tab:model_configure}
\resizebox{\textwidth}{!}{%
\begin{tabular}{l ccc}
\toprule
\textbf{Setting}
& \textbf{Qwen3-1.7B}
& \textbf{Qwen3-4B / -8B}
& \textbf{Gemma-4-E4B} \\
\midrule
Device
& 1 node, 8 GPUs
& 4 nodes, 24+8 GPUs
& 4 nodes, 24+8 GPUs \\
Student
& Qwen3-1.7B
& Qwen3-4B, Qwen3-8B
& Gemma-4-E4B-it \\
Teacher (non-thinking)
& Qwen3-4B-Instruct-2507
& Qwen3-30B-A3B-Instruct-2507
& --- \\
Teacher (thinking)
& Qwen3-4B-Thinking-2507
& Qwen3-30B-A3B-Thinking-2507
& Gemma-4-31B-it \\
Teacher-base
& Qwen3-4B-Base
& Qwen3-30B-A3B-Base
& Gemma-4-31B \\
Thinking mode
& non-thinking + thinking
& non-thinking + thinking
& thinking \\
vLLM rollout mode
& colocate (TP=4)
& server node (TP=8)
& server node (TP=8) \\
Effective global batch
& 256
& 288
& 288 \\
\bottomrule
\end{tabular}%
}
\end{table}
\begin{table}[h]
\centering
\small
\setlength{\tabcolsep}{7pt}
\caption{\textbf{Common training parameters.} Training parameters shared across all models.}
\label{tab:common_configure}
\begin{tabular}{@{}lc@{}}
\toprule
\textbf{Setting} & \textbf{Value} \\
\midrule
Learning rate        & $5\times10^{-6}$ \\
LR schedule          & cosine decay \\
Minimum LR ratio     & 0.1 \\
Warmup ratio         & 0.1 \\
Optimizer             & AdamW \\
Gradient clipping     & 1.0 \\
Training steps        & 100 \\
Maximum generation length & 8192 \\
Sampling temperature & 0.7 \\
KL coefficient $\beta$ & 0.0 \\
Gradient multiplier & 0.1 \\
\bottomrule
\end{tabular}
\end{table}

\newpage
Table~\ref{tab:eval_config} summarizes the evaluation configurations for all benchmarks. We report pass@1, interpreted as single-sample accuracy, and average the scores over multiple independent repetitions to reduce the variance introduced by stochastic generation. The repetitions are not used for best-of-$k$ selection. AIME24 and AIME25 are evaluated with 16 repetitions because of their relatively small number of problems, whereas larger benchmarks are evaluated with fewer repetitions.

\begin{table}[h]
\centering
\small
\setlength{\tabcolsep}{6pt}
\caption{\textbf{Evaluation benchmark configs.} For each benchmark, we report pass@1 (single-sample accuracy), averaged across the listed number of repetitions.}
\label{tab:eval_config}
\begin{tabular}{llc}
\toprule
\textbf{Domain} & \textbf{Benchmark} & \textbf{Repetitions} \\
\midrule
\multirow{7}{*}{Math}
 & AIME24        & 16 \\
 & AIME25        & 16 \\
 & AMC23         & 10 \\
 & HMMT          & 10 \\
 & MATH500       & 10 \\
 & OlympiadBench & 3  \\
 & RGMath        & 3  \\
\midrule
\multirow{4}{*}{Code}
 & LiveCodeBenchv5 & 3 \\
 & CodeForces      & 3 \\
 & CodeContests    & 3 \\
 & RGAlgorithmic   & 3 \\
\midrule
\multirow{3}{*}{Science}
 & GPQADiamond & 10 \\
 & SuperGPQA   & 3  \\
 & SciBench    & 3  \\
\bottomrule
\end{tabular}
\end{table}

\end{document}